\documentclass[11pt,a4paper]{article}
\usepackage{times,latexsym}
\usepackage{url}
\usepackage[T1]{fontenc}
\usepackage[english]{babel}
\usepackage[table]{xcolor}
\usepackage{collcell}
\usepackage{hhline}
\usepackage{pgf}
\usepackage{booktabs}
\usepackage{multirow}
\usepackage{arydshln}
\usepackage{linguex}
\alignSubExtrue 
\usepackage[acceptedWithA]{tacl2018v2} 

\usepackage{extdash} 

\usepackage{hhline}
\usepackage{amssymb}
\usepackage{multirow}
\usepackage{arydshln}
\usepackage{enumitem}
\usepackage{multicol}

\usepackage{xspace,mfirstuc,tabulary}

\newif\iftaclinstructions
\taclinstructionsfalse 
\iftaclinstructions
\renewcommand{\confidential}{}
\renewcommand{\anonsubtext}{(No author info supplied here, for consistency with
TACL-submission anonymization requirements)}
\newcommand{\instr}
\fi

\iftaclpubformat 

\else

\fi

\def\colorModel{hsb}
\newcommand\ColCell[1]{
  \pgfmathparse{#1<10?1:0}  
    \ifnum\pgfmathresult=0\relax\color{white}\fi
    \pgfmathsetmacro\compA{135/360} 
    \pgfmathsetmacro\compB{1}       
    \pgfmathsetmacro\compC{(#1-25)/100}            
  \edef\x{\noexpand\centering\noexpand\cellcolor[\colorModel]{\compA,\compB,\compC}}\x #1
  } 
\newcolumntype{E}{>{\collectcell\ColCell}m{4.5ex}<{\endcollectcell}}  

\newcommand\ColCellTwo[1]{
  \pgfmathparse{#1>10?1:0}  
    \ifnum\pgfmathresult=0\relax\color{white}\fi
    \pgfmathsetmacro\compA{200/360} 
    \pgfmathsetmacro\compB{1}       
    \pgfmathsetmacro\compC{(1.3-#1)}            
  \edef\x{\noexpand\centering\noexpand\cellcolor[\colorModel]{\compA,\compB,\compC}}\x #1
  } 
\newcolumntype{F}{>{\collectcell\ColCellTwo}m{4.5ex}<{\endcollectcell}}  

\newcommand\ColCellThree[1]{
  \pgfmathparse{#1<50?1:0}  
    \relax\color{black}
    \ifnum\pgfmathresult=0\pgfmathsetmacro\compA{135/360}\fi 
    \ifnum\pgfmathresult=1\pgfmathsetmacro\compA{10/360}\fi 
    \ifnum\pgfmathresult=0\pgfmathsetmacro\compB{(#1-50)/60}\fi
    \ifnum\pgfmathresult=1\pgfmathsetmacro\compB{(50-#1)/60}\fi 
    \pgfmathsetmacro\compC{1}
  \edef\x{\noexpand\centering\noexpand\cellcolor[\colorModel]{\compA,\compB,\compC}}\x #1
  } 
\newcolumntype{G}{>{\collectcell\ColCellThree}m{1.5ex}<{\endcollectcell}}  

\usepackage{graphicx,scalerel} 
\graphicspath{{./images/}} 
\usepackage{adjustbox}
\usepackage{graphicx}

\def\BLiMP{BLiMP}

\def\newterm#1{\textit{#1}}

\def\cites#1{\citeauthor{#1}'s (\citeyear{#1})}
\newcolumntype{P}[1]{>{\arraybackslash}p{#1}}

\title{BLiMP: The Benchmark of Linguistic Minimal Pairs for English}

\author{Alex Warstadt$^1$, 
        Alicia Parrish$^1$, 
        Haokun Liu$^2$,
        Anhad Mohananey$^2$,\\\bf 
        Wei Peng$^2$, 
        Sheng-Fu Wang$^1$, 
        Samuel R. Bowman$^{1,2,3}$ \\\AND
\textnormal{$^1$Dept. of Linguistics}\\ New York University \And
\textnormal{$^2$Dept. of Computer Science}\\ New York University\And
\textnormal{$^3$Center for Data Science}\\ New York University \AND Correspondence: \href{mailto:warstadt@nyu.edu}{\tt warstadt@nyu.edu}}

\date{}

\begin{document}
\maketitle
\begin{abstract}

We introduce The Benchmark of Linguistic Minimal Pairs (BLiMP),\footnote{\href{https://github.com/alexwarstadt/blimp}{\url{https://github.com/alexwarstadt/blimp}}} a challenge set for evaluating the linguistic knowledge of language models (LMs) on major grammatical phenomena in English. \BLiMP{} consists of 67 individual datasets, each containing 1,000 minimal pairs, i.e.\ pairs of minimally different sentences that contrast in grammatical acceptability and isolate specific phenomenon in syntax, morphology, or semantics. We generate the data according to linguist-crafted grammar templates, and human aggregate agreement with the labels is 96.4\%. We evaluate $n$-gram, LSTM, and Transformer (GPT-2 and Transformer-XL) LMs by observing whether they assign a higher probability to the acceptable sentence in each minimal pair. We find that state-of-the-art models identify morphological contrasts related to agreement reliably, but they struggle with some subtle semantic and syntactic phenomena, such as negative polarity items and extraction islands.

\end{abstract}

\section{Introduction} 

Current neural networks for sentence processing rely on unsupervised pretraining tasks like language modeling. Still, it is an open question how the linguistic knowledge of state-of-the-art language models (LMs) varies across the linguistic phenomena of English. Recent studies \citep[e.g.,][]{linzen2016assessing,marvin2018targeted,wilcox2018filler} have explored this question by evaluating LMs' preferences between \newterm{minimal pairs} of sentences differing in grammatical acceptability as in Example \ref{eg:minimal}. However, each of these studies uses a different set of metrics, and focuses on a small set of linguistic paradigms, severely limiting any possible big-picture conclusions.

\ex.\label{eg:minimal} 
\a. The cats annoy Tim. (\textit{grammatical})
\b. * The cats annoys Tim. (\textit{ungrammatical})

We introduce the Benchmark of Linguistic Minimal Pairs (shortened to BLiMP), a linguistically-motivated benchmark for assessing the sensitivity of LMs to acceptability contrasts across a wide range of English phenomena, covering both previously-studied and novel contrasts. \BLiMP{} consists of 67 datasets automatically generated from linguist-crafted grammar templates, each containing 1,000 minimal pairs and organized by phenomenon into 12 categories. Validation with crowd workers shows that \BLiMP{} faithfully represents human preferences. 

    \begin{table*}[t]\small\centering
    \begin{tabular}{llp{0.34\linewidth}p{0.34\linewidth}}
    \toprule
    Phenomenon & N & Acceptable Example & Unacceptable Example\\
    \midrule
    \textsc{Anaphor agr.} & 2 & \emph{Many girls insulted \underline{themselves}.} & \emph{Many girls insulted \underline{herself}.}\\
    \textsc{Arg. structure} & 9 & \emph{Rose wasn't \underline{disturbing} Mark.} & \emph{Rose wasn't \underline{boasting} Mark.}\\
    \textsc{Binding} & 7 & \emph{Carlos said that Lori helped \underline{him}.} & \emph{Carlos said that Lori helped  \underline{himself}.} \\
    \textsc{Control/raising} & 5 & \emph{There was \underline{bound} to be a fish escaping.} & \emph{There was \underline{unable} to be a fish escaping.}\\
    \textsc{Det.-noun agr.} & 8 & \emph{Rachelle had bought that \underline{chair}.} & \emph{Rachelle had bought that \underline{chairs}.}\\
    \textsc{Ellipsis} & 2 & \emph{Anne's doctor cleans one \underline{important} \phantom{xxx} book and Stacey cleans a few.} & \emph{Anne's doctor cleans one book and \phantom{xxx} \phantom{xxx} Stacey cleans a few \underline{important}.}\\
    \textsc{Filler-gap} & 7 & \emph{Brett knew \underline{what} many waiters find.} & \emph{Brett knew \underline{that} many waiters find.}\\
    \textsc{Irregular forms} & 2 & \emph{Aaron \underline{broke} the unicycle.} & \emph{Aaron \underline{broken} the unicycle.}\\
    \textsc{Island effects} & 8 & \emph{Which \underline{bikes} is John fixing?} & \emph{Which is John fixing \underline{bikes}?}\\
    \textsc{NPI licensing} & 7 & \emph{The truck has \underline{clearly} tipped over.} & \emph{The truck has \underline{ever} tipped over.}\\
    \textsc{Quantifiers} & 4 & \emph{No boy knew \underline{fewer than} six guys.} & \emph{No boy knew \underline{at most} six guys.}\\
    \textsc{Subject-verb agr.} & 6 & \emph{These casseroles \underline{disgust} Kayla.} & \emph{These casseroles \underline{disgusts} Kayla.}\\
    \bottomrule
    \end{tabular}
    \caption{Minimal pairs from each of the twelve linguistic phenomenon categories covered by \BLiMP. Differences are underlined. \textit{N} is the number of 1,000-example minimal pair paradigms within each broad category.}\label{tab:phenomena table}
    \end{table*}

We use \BLiMP{} to study several pretrained LMs: Transformer-based LMs GPT-2 \cite{radford2019language} and Transformer-XL \cite{dai2019transformer}, an LSTM LM trained by \citet{gulordava2019colorless}, and an $n$-gram LM. We evaluate whether the LM assigns a higher probability to the acceptable sentence in each minimal pair to determine which grammatical distinctions LMs are sensitive to. This gives us indirect evidence about each model's linguistic knowledge and allows us to compare models in a fine-grained way. We conclude that current neural LMs appear to acquire robust knowledge of morphological agreement and some syntactic phenomena such as ellipsis and control/raising. They show weaker evidence of knowledge about argument structure, negative polarity item licensing, and the semantic properties of quantifiers. All models perform at or near chance on extraction islands. Overall, every model we evaluate falls short of human performance by a wide margin. GPT-2, which performs the best, performs 8 points below humans overall, though it does match or exceed human performance on specific phenomena.

In \S \ref{sec:data_learning_curve_experiment} we conduct additional experiments to investigate the effect of training size on the LSTM LM and Transformer-XL's performance on \BLiMP{}. While we see steady improvements in overall performance, we find that LMs learn phenomenon-specific distinctions at different rates. In \S \ref{sec:prefix} we consider alternative well-motivated evaluation metrics on \BLiMP{}, but find that they do not differ drastically from our method of comparing LM probabilities for full sentences.

We conclude that while models like GPT-2 appear to have significant linguistic knowledge, this knowledge is concentrated in some specific domains of English grammar. We use \BLiMP{} to uncover several linguistic phenomena where even state-of-the-art language models clearly lack human-like knowledge, and to bring into focus those areas of grammar that future studies evaluating LMs should investigate in greater depth.

\section{Background \& Related Work} 

\subsection{Language Models} 
    The objective of a language model is to give a probability distribution over the strings of a language. Both neural network and non-neural network architectures are used to build LMs, and neural models can be trained in a \newterm{self-supervised} setting without the need for labeled data. Recently, variants of neural language modeling have been shown to be a strong pretraining task for natural language processing tasks \cite{howard2018universal, peters2018deep, radford2018improving,devlin2019bert}.
    
    The last decade has seen two major paradigm shifts in the state of the art for language modeling. First, there was a movement from  models based on local $n$-gram statistics \citep[see][]{chen1999empirical} to neural sequence models such as LSTMs \cite{mikolov2010recurrent}, which optimize on the task of predicting the next token. Subsequently, Transformer-based architectures employing self-attention \cite{vaswani2017attention} have outperformed LSTMs \citep[e.g.,][]{dai2019transformer}. Although these shifts have resulted in stronger LMs, perplexity on large benchmark datasets like WikiText-103 \citep{merity16pointer} has remained the primary performance metric, which cannot give detailed insight into these models' knowledge of grammar. Evaluation on benchmarks like GLUE \cite{wang2018glue,wang2019superglue}, that heavily adapt language models to perform downstream tasks, is more informative, but doesn't offer broad coverage of linguistic phenomena, and doesn't necessary  reflect knowledge that is already present in the LMs.

\begin{table*}[]
    \centering
    \begin{tabular}{lp{0.74\textwidth}}
    \toprule
        Phenomenon & Relevant work \\\midrule
        Anaphora/binding & 
            \citet{marvin2018targeted}, 
            \citet{futrell2018psycholinguistic}, 
            \citet{warstadt2019neural} 
            \\
        Subj.-verb agreement & 
            \citet{linzen2016assessing}, 
            \citet{futrell2018psycholinguistic}, 
            \citet{gulordava2019colorless},
            \citet{marvin2018targeted}, 
            \citet{an2019representation},
            \citet{warstadt2019neural}
            \\
        Neg. polarity items &  
            \citet{marvin2018targeted},  
            \citet{futrell2018psycholinguistic}, 
            \citet{jumelet2018language},
            \citet{wilcox2019structural},
            \citet{warstadt2019investigating}
            \\
        Filler-gap/Islands & 
            \citet{wilcox2018filler}, 
            \citet{warstadt2019neural},
            \citet{chowdhury2018rnn,chowdhury2019lstm}
            \citet{chaves2020dont}, 
            \citet{dacosta2020assessing}
            \\
        Argument structure & 
            \citet{kann2019verb},
            \citet{warstadt2019neural}, 
            \citet{chowdhury2019lstm}
            \\
        \bottomrule
    \end{tabular}
    \caption{Summary of related work organized by linguistic phenomena tested. All studies analyze neural networks using acceptability judgments on minimal pairs mainly in English. Some studies appear multiple times.}
    \label{tab:related_work}
\end{table*}

\subsection{Linguistic Knowledge of NNs}

Many recent studies have searched for evidence that neural networks (NNs) learn representations that implicitly encode grammatical concepts. We refer to the ability to encode these concepts as \newterm{linguistic knowledge}. Some studies evaluate NNs' linguistic knowledge using probing tasks in which a classifier is trained to directly predict grammatical properties of a sentence (e.g.~syntactic tree depth) or part of a sentence (e.g.~part-of-speech) using only the NNs' learned representation as input \citep{shi2016syntax,adi2017fine,conneau2018cram,ettinger2018assessing,tenney2019you}. We follow a complementary approach that uses acceptability judgments to address the same question without the need for training data labeled with grammatical concepts. Acceptability judgments are the main form of behavioral data used in generative linguistics to measure human linguistic competence \citep{chomsky1965aspects,schutze1996empirical}. 

One branch of this literature uses minimal pairs to infer whether LMs detect specific grammatical contrasts. Table \ref{tab:related_work} summarizes linguistic phenomena studied in this work. For instance, \newcite{linzen2016assessing} look closely at minimal pairs contrasting subject-verb agreement. \newcite{marvin2018targeted} expand the investigation to negative polarity item and reflexive licensing. However, these and related studies cover a limited set of phenomena, to the exclusion of well-studied phenomena in linguistics such as control and raising, ellipsis, quantification, and countless others. This is likely due to the labor-intensive nature of collecting such targeted minimal pairs.


A related line of work evaluates neural networks on acceptability judgments in a more domain-general way. Corpora of sentences and their grammaticality are collected for this purpose in a number of studies \citep{heilman2014predicting, lau2017grammaticality, warstadt2019neural}. The most recent and comprehensive corpus is CoLA \citep{warstadt2019neural}, containing 10k sentences covering a wide variety of linguistic phenomena provided as examples in linguistics papers and books. CoLA, which is included in the GLUE benchmark \citep{wang2018glue}, has been used to track advances in the sensitivity of reusable sentence encoding models to acceptability. Current models like BERT \citep{devlin2019bert} and T5 \cite{raffel2019t5} now learn to give acceptability judgments that approach or even exceed individual human agreement with CoLA.

While CoLA can provide evidence about phenomenon-specific knowledge of models, this method is limited by the need to train a supervised classifier on CoLA data prior to evaluation. This is because CoLA is designed for binary acceptability classification, and there is no generally accepted method for obtaining binary acceptability predictions from unsupervised models like LMs.\footnote{Though see \citet{lau2017grammaticality} for some promising proposals for normalizing LM probabilities to correlate with gradient acceptability.} \citet{warstadt2019grammatical} measure phenomenon-specific performance on CoLA for several pretrained sentence encoding models: an LSTM, GPT \citep{radford2018improving}, and BERT. 
However, the use of supervision prevents making strong conclusions about the sentence encoding component, since it is not possible to distinguish what the encoder knows from what is learned through supervised training on acceptability data.

Evaluating LMs on minimal pairs avoids this problem, with the caveat that the LM probability of a sentence can only serve as a proxy for acceptability if confounding factors impacting a sentence's probability such as length and lexical content are controlled for. It is with these considerations in mind that we design \BLiMP{}.

\section{Data}\label{sec:data} 
\BLiMP{} 
consists of 67 minimal pair paradigms, each with 1,000 sentence pairs in mainstream American English grouped into 12 categories.\footnote{We choose English because it is the native language of the linguists who built the grammar templates, though in the long run, creating versions of \BLiMP{} in additional languages would allow for coverage of more phenomena and expand BLiMP's range of usefulness. We assume 1000 pairs is sufficient to limit random noise resulting from small sample sizes.} We refer to minimal pair types as \newterm{paradigms} and categories as \newterm{phenomena}. Each paradigm is annotated for the unique contrast it isolates and the broader phenomena it is part of. We automatically generate the data from linguist-crafted grammar templates, and our automatic labels are validated with crowd-sourced human judgments. 

While each minimal pair type corresponds to exactly one paradigm, a particular fact about English grammar may be illustrated by multiple paradigms. For instance, the fact that certain determiners and nouns agree can be illustrated by keeping the determiner the same and changing the number marking of the noun as in the example in Table \ref{tab:phenomena table}, or by keeping the noun the same and changing the determiner (e.g. \emph{Rachelle had bought those chair.}). With completeness in mind, we include such complementary paradigms in \BLiMP{} whenever possible.
 

\subsection{Data generation procedure}
To create minimal pairs exemplifying a wide array of linguistic contrasts, we found it necessary to artificially generate all datasets. This ensures both that we have sufficient unacceptable examples, and that the data is fully controlled, allowing for repeated isolation of a single linguistic phenomenon \citep{ettinger2018assessing}. For each paradigm, we use a generation script to sample lexical items from a vocabulary of over 3,000 items according to a template specifying linear order of the phrases in the acceptable and unacceptable sentences in each minimal pair. Our data generation scripts are publicly available.\footnote{\href{https://github.com/alexwarstadt/data_generation}{\url{https://github.com/alexwarstadt/data_generation}}} We annotate these lexical items with the morphological, syntactic, and semantic features needed to enforce selectional restrictions and create grammatical and semantically felicitous sentences.

All examples in a paradigm are structurally analogous up to the point required for the relevant contrast but may vary in some ways. For instance, the template for \textsc{NPI licensing}, illustrated in Table \ref{tab:phenomena table}, specifies that an arbitrary verb phrase needs to be generated. Accordingly, the generation script samples from the entire set of verbs and generates the required arguments on-the-fly. Thus, the structure of the sentence then depends on whether the sampled verb is transitive, clause-embedding, raising, etc., but that same verb phrase and its arguments are used in both pairs in the paradigm.

    
    

This generation procedure is not without limitations, and despite the very detailed vocabulary we use, implausible sentences are occasionally generated (e.g., \emph{Sam ran around some glaciers}). In these cases, though, both the acceptable and unacceptable sentences will be equally implausible given world knowledge, so any difference in the probability assigned to them is still attributable to the intended grammatical contrast.

\subsection{Coverage}
    
The paradigms covered by \BLiMP{} represent well-established contrasts in English morphology, syntax, and semantics. Each paradigm is grouped into one of 12 phenomena, shown in Table \ref{tab:phenomena table}. Examples of all 67 paradigms appear in Table \ref{tab:appendix} of the Appendix.
The paradigms are selected with the constraints that they can be characterized using templates as described above and illustrated with minimal pairs of sentences equal in length\footnote{We define length as the number of entries from our lexicon. Some sentences in a pair contain different numbers of words because \emph{visit} and \emph{drop by} are each one lexical entry. Where discrepancies in number of words occur, they are generally randomly distributed across the grammatical and ungrammatical sentences in a paradigm.} that differ in at most one vocabulary item.

While this dataset has broad coverage, it is not exhaustive. It is not possible to include every grammatical phenomenon of English, and there is no agreed-upon set of core phenomena. However, we consider frequent inclusion of a phenomenon in a syntax/semantics textbook as an informal proxy for what linguists consider to be core phenomena. We survey several syntax textbooks \citep[e.g.,][]{sag2003syntactic,adger2003core,sportiche2013introduction}, and find that nearly all of the phenomena in \BLiMP{} are discussed in some source. Most of the topics that repeatedly appear in textbooks and can be represented with minimal pairs (e.g. agreement, control/raising, wh-extraction/islands, binding) are present in \BLiMP{}.\footnote{In line with these textbooks, we rely on stereotyped gender-name pairings and contrasts not present in all English dialects (more detail provided in the appendix).}

We characterize the 12 phenomena in \BLiMP{} as follows\footnote{Our implementation of these phenomena is often narrower than the linguistic definition due to the particular constraints described above.}: 
\begin{itemize}[noitemsep,leftmargin=*,topsep=0pt]
    \item \textsc{Anaphor agreement}: the requirement that reflexive pronouns like \emph{himself} (a.k.a.~anaphora) agree with their antecedents in person, number, gender and animacy.
    \item \textsc{Argument structure}: the ability of different verbs to appear with different types of arguments. For instance, different verbs can appear with a direct object, participate in the causative alternation, or take an inanimate argument.
    \item \textsc{Binding}: the structural relationship between a pronoun and its antecedent. All paradigms illustrate aspects of \cites{chomsky1981lectures} Principle A. Since coindexation cannot be annotated in \BLiMP{}, Principles B and C are not illustrated. 
    \item \textsc{Control/raising}: syntactic and semantic differences between various types of predicates that embed an infinitival VP. This includes control, raising, and \emph{tough}-movement predicates.
    \item \textsc{Determiner-noun agreement}: number agreement between demonstrative determiners (e.g.~\emph{this}/\emph{these}) and the associated noun. 
    \item \textsc{Ellipsis}: the possibility of omitting expressions from a sentence. Since this is difficult to illustrate with sentences of equal length, our paradigms cover only special cases of noun phrase ellipsis that meet this constraint. 
    \item \textsc{Filler-gap}: dependencies arising from phrasal movement in, e.g., \emph{wh}-questions. 
    \item \textsc{Irregular forms}: irregular morphology on English past participles (e.g. \emph{broken}). We are unable to evaluate models on non-existent forms like \emph{*breaked} because such forms are out of the vocabulary for some LMs. 
    \item \textsc{Island effects}: restrictions on syntactic environments where the gap in a filler-gap dependency may occur.
    \item \textsc{NPI licensing}: restrictions on the distribution of \newterm{negative polarity items} like \emph{any} and \emph{ever} limited to, e.g., the scope of negation and \emph{only}. 
    \item \textsc{Quantifiers}: restrictions on the distribution of quantifiers. We cover two such restrictions: superlative quantifiers (e.g., \emph{at least}) cannot embed under negation, and definite quantifiers and determiners cannot be subjects in existential-\emph{there} constructions. 
    \item \textsc{Subject-verb agreement}: subjects and present tense verbs must agree in number.
\end{itemize}


\subsection{Comparison to Related Resources}
With a vocabulary of over 3,000 words, \BLiMP{} has by far the most lexical variation of any related generated dataset. It includes verbs with 11 different subcategorization frames, including verbs that select for PPs, infinitival VPs, and embedded clauses. By comparison, datasets by \citet{ettinger2018assessing} and \citet{marvin2018targeted} use vocabularies of under 200 items. Other datasets of minimal pairs that achieve more lexical and syntactic variety use data-creation methods that limit empirical scope and control. \citet{linzen2016assessing} construct a dataset of minimal pairs for subject-verb agreement by changing verbs' number marking in a subset of English Wikipedia, but this approach does not generalize beyond agreement phenomena. \citet{lau2017grammaticality} construct minimal pairs by taking sentences from the BNC through round-trip machine translation. The resulting sentences contain a wider variety of grammatical violations, but it is not possible to control the nature or quantity of violations in the resulting sentences.

    
    \subsection{Data validation}\label{sec:data_validation}
    To verify that the generated sentences represent a real contrast in acceptability, we conduct human validation via Amazon Mechanical Turk.\footnote{The full set of human judgments and a summary of the results for all 67 paradigms is in Table \ref{tab:appendix} in the Appendix.}
    Twenty separate validators rated five pairs from each of the 67 paradigms, for a total of 6700 judgments.
    We restricted validators to individuals currently located in the US who self-reported as native speakers of English.
    To assure that our validators made a genuine effort on the task, each HIT included an attention check item and a hidden field question to catch bot-assisted humans. 
    Validators were paid \$0.25 for completing 5 judgments, which we estimate took 1-2 minutes.
    For each minimal pair, 20 individuals completed a forced-choice task mirroring the LMs' task; the human-determined acceptable sentence was calculated via majority vote of annotators. By this metric, we estimate aggregate human agreement with our annotations to be 96.4\% overall. As a threshold of inclusion in \BLiMP{}, the majority of validators needed to agree with \BLiMP\ on at least 4/5 examples from each paradigm. Thus, all 67 paradigms in the public version of \BLiMP{} passed this validation; only two additional paradigms were rejected on this criterion.
    We also estimate \emph{individual} human agreement to be 88.6\% overall using the approximately 100 annotations from each paradigm.\footnote{A few had to be excluded due to ineligible annotators.} Figure \ref{tab:results table} reports individual human results (and model results) as a conservative measure of human agreement.

\newcommand\items{13}   
\newcommand\rotation{30} 
\arrayrulecolor{white} 
\begin{table*}\small\centering
\begin{adjustbox}{max width=\textwidth}
\renewcommand{\arraystretch}{1.1}
\begin{tabular}{l*{\items}{|E}|}
\multicolumn{1}{p{2.5ex}}{\rotatebox{0}{Model}} &
\multicolumn{1}{p{2.5ex}}{\rotatebox{\rotation}{\textbf{Overall}}} &
\multicolumn{1}{p{2.5ex}}{\rotatebox{\rotation}{\textsc{Ana.\ agr}}} &
\multicolumn{1}{p{2.5ex}}{\rotatebox{\rotation}{\textsc{Arg.\ str}}} &
\multicolumn{1}{p{2.5ex}}{\rotatebox{\rotation}{\textsc{Binding}}} &
\multicolumn{1}{p{2.5ex}}{\rotatebox{\rotation}{\textsc{Ctrl.\ rais.}}} &
\multicolumn{1}{p{2.5ex}}{\rotatebox{\rotation}{\textsc{D-n agr}}} &
\multicolumn{1}{p{2.5ex}}{\rotatebox{\rotation}{\textsc{Ellipsis}}} &
\multicolumn{1}{p{2.5ex}}{\rotatebox{\rotation}{\textsc{Filler.\ gap}}} &
\multicolumn{1}{p{2.5ex}}{\rotatebox{\rotation}{\textsc{Irregular}}} &
\multicolumn{1}{p{2.5ex}}{\rotatebox{\rotation}{\textsc{Island}}} &
\multicolumn{1}{p{2.5ex}}{\rotatebox{\rotation}{\textsc{NPI}}} &
\multicolumn{1}{p{2.5ex}}{\rotatebox{\rotation}{\textsc{Quantifiers}}} &
\multicolumn{1}{p{2.5ex}}{\rotatebox{\rotation}{\textsc{S-v agr}}} \\ \hhline{~*\items{|-}|}
5-gram & 61.2 & 47.9 & 71.9 & 64.4 & 68.5 & 70.0 & 36.9 & 60.2 & 79.5 & 57.2 & 45.5 & 53.5 & 60.3\\ \hhline{~*\items{|-}|}
LSTM & 69.8 & 91.7  & 73.2 & 73.5 & 67.0 & 85.4 & 67.6 & 73.9 & 89.1 & 46.6 & 51.7 & 64.5 & 80.1 \\ \hhline{~*\items{|-}|}
TXL & 69.6 & 94.1  & 72.2 & 74.7 & 71.5 & 83.0 & 77.2 & 66.6 & 78.2 & 48.4 & 55.2 & 69.3 & 76.0 \\ \hhline{~*\items{|-}|}
GPT-2  & 83.0 & 99.3  & 81.8 & 80.9 & 81.9 & 95.8 & 89.3 & 81.3 & 91.9 & 72.7 & 76.8 & 79.0 & 86.4\\ \hhline{~*\items{|-}|}
Human & 88.6 & 97.5 & 90.0 & 87.3 & 83.9 & 92.2 & 85.0 & 86.9 & 97.0 & 84.9 & 88.1 & 86.6 & 90.9 \\ \hhline{~*\items{|-}|}
\end{tabular}
\end{adjustbox}
\caption{Percentage accuracy of four baseline models and raw human performance on \BLiMP\ using a forced-choice task. A random guessing baseline would achieve an accuracy of 50\%.}\label{tab:results table}
\end{table*}


\section{Models} 

    \paragraph{GPT-2} 
    GPT-2 \citep{radford2019language} is a large-scale language model using the Transformer architecture \cite{vaswani2017attention}. 
    Our main experiments use GPT-2-large with 36 layers and 774M parameters.\footnote{GPT-2-XL performs slightly worse on \BLiMP{}; see \S \ref{sec:data_learning_curve_experiment}.} The model is pretrained on \citeauthor{radford2019language}'s WebText dataset, which contains 40GB of English text extracted from web pages and filtered for quality. To our knowledge, WebText is not publicly available, so assuming an average of 5-6 bytes/chars per word, we estimate WebText contains about 8B tokens.
    We use \texttt{jiant}, a codebase for training and evaluating sentence understanding models \cite{wang2019jiant}, to implement code for evaluating GPT-2 on \BLiMP{}.\footnote{\href{https://github.com/nyu-mll/jiant/tree/blimp-and-npi/scripts/blimp}{\url{https://github.com/nyu-mll/jiant/tree/blimp-and-npi/scripts/blimp}}}
    
    \paragraph{Transformer-XL} 
    Transformer-XL \cite{dai2019transformer} is another multi-layer Transformer-based neural language model.
    We test the pretrained Transformer-XL Large model with 18 layers of Transformer decoders and 16 attention heads for each layer. The model is trained on WikiText-103 \cite{merity16pointer}, a corpus of 103M tokens from English Wikipedia. 
    Code for testing Transformer-XL on \BLiMP{} is also implemented in \texttt{jiant}.

    \paragraph{LSTM} 
    We include a long-short term memory (LSTM, \citealp{hochreiter1993long}) LM in our experiments. Specifically, we test a pretrained LSTM LM from \cite{gulordava2019colorless} on \BLiMP{}. The model is trained on a 83M-token corpus extracted from English Wikipedia. To investigate the effect of training size on model performance (\S\ref{sec:data_learning_curve_experiment}), we retrain a series of LSTM and Transformer-XL models with the same hyperparameters and the following training sizes: 64M, 32M, 16M, 8M, 4M, 2M, 1M, 1/2M, 1/4M, and 1/8M tokens. For each size, we train the model on five different random samples of the original training data, which has a size of 83M tokens.\footnote{\href{https://github.com/sheng-fu/colorlessgreenRNNs}{\url{https://github.com/sheng-fu/colorlessgreenRNNs}}}

    \paragraph{5-gram} 
    We build a 5-gram LM on the English Gigaword corpus \cite{graff2003english}, which consists of 3.1B tokens.
    To efficiently query $n$-grams we use an implementation\footnote{\href{https://github.com/kpu/kenlm}{\url{https://github.com/kpu/kenlm}}} based on \citet{heafield2013scalable}.\footnote{\href{https://github.com/anhad13/blimp_ngram}{\url{https://github.com/anhad13/blimp_ngram}}}


\section{Results \& Discussion}\label{sec:results}

An LM's overall accuracy on \BLiMP{} is simply the proportion of the 67,000 minimal pairs in which the model assigns a higher probability to the acceptable sentence. We report the results for all models and human evaluation in Table \ref{tab:results table}. GPT-2 achieves the highest accuracy and the 5-gram model the lowest. 
All models perform well below estimated human accuracy (as described in \S \ref{sec:data_validation}). The 5-gram model's poor performance---overall and on every individual category---indicates that \BLiMP{} is likely not solvable from local co-occurrence statistics alone.  

Because we evaluate pretrained models that differ in architecture and training data, we can only speculate about what drives these differences (though see \S \ref{sec:data_learning_curve_experiment} for a controlled ablation study on the LSTM LM). The results seem to indicate that access to training data is the main driver of performance on \BLiMP{} for the neural models we evaluate. This may explain why Transformer-XL and the LSTM LM perform similarly in spite of differences in architecture, as both are trained on approximately $100$M tokens of Wikipedia text. Relatedly, GPT-2's advantage may come from the fact that it is trained on roughly two orders of magnitude more data. Possibly, LSTMs trained on larger datasets could perform comparably to GPT-2, but such experiments are impractical due to the inefficiency of training LSTMs at this scale.

\subsection{Results \& Discussion by Phenomenon}
The results also give insight into how LM's linguistic knowledge varies by domain. Models generally perform best and closest to human level on morphological phenomena. For instance, GPT-2 performs within 5 points of humans on \textsc{anaphor agr.}, \textsc{det.-noun agr.}, and \textsc{subj.-verb agr}. The set of challenging phenomena is more diverse. \textsc{Islands} are the hardest phenomenon for most models. Only GPT-2 performs well above chance, and it remains 12 points below humans. Some semantic phenomena, specifically those involving \textsc{NPI licensing} and \textsc{quantifiers}, are also challenging overall. All models perform relatively poorly on \textsc{arg. structure}.

From these results we conclude that current SotA LMs robustly encode basic facts of English agreement. This does not mean that LMs will come close to human performance for all agreement phenomena. \S \ref{sec:long-distance} discusses evidence that increased dependency length and the presence of agreement attractors of the kind investigated by \citet{linzen2016assessing} and \citet{gulordava2019colorless} reduce performance on agreement phenomena.

We find, in accordance with \citet{wilcox2018filler}, that LMs do represent long-distance wh-dependencies, but we also conclude that their representations differ fundamentally from humans'. While some models approach human performance in ordinary filler-gap dependencies, they are exceptionally poor at identifying island violations overall. This finding suggests that they reliably encode long-distance dependencies in general, but not the syntactic domains in which these dependencies are blocked, though GPT-2 does perform well above chance on some paradigms of \textsc{island effects}. However, strong conclusions about how these models represent wh-dependencies are not possible using the forced-choice task compatible with \BLiMP{}, and a complete assessment of syntactic islands is best addressed using a factorial design that manipulates both the presence of an island and an attempt to extract from it as in \citet{kush2018investigating} or \citet{wilcox2018filler}. 

In the semantic phenomena where models struggle (\textsc{NPIs} and \textsc{Quantifiers}), violations are often attributed in semantic theories to a presupposition failure or contradiction arising from semantic composition or pragmatic reasoning \citep[e.g.,][]{chierchia2013logic,ward1995definiteness,geurts2007least}. These abstract semantic and pragmatic factors may be difficult for LMs to learn. \citeauthor{marvin2018targeted} also find that LSTMs largely fail to recognize NPI licensing conditions. \citet{warstadt2019investigating} find that BERT (which is similar in scale to GPT-2) recognizes these conditions inconsistently in an unsupervised setting. 

The weak performance on \textsc{arg. structure} is somewhat surprising, since arguments and heads are usually---though not always---adjacent (e.g., subjects and direct objects are adjacent to the verb in default English word order). However, argument structure is closely related to semantic event structure \citep[see][]{marantz2013verbal}, which may be comparatively difficult for LMs to learn. Also, judgments about argument structure are complicated by the possibility of coercing a frequently transitive verb to be intransitive and vice versa as well as the existence of secondary meanings of verbs with different argument structures (e.g.~normally intransitive \emph{boast} has a transitive use as in \emph{The spa boasts 10 pools}), which might make this domain somewhat more difficult for LMs. Though even with these complications, humans detect the intended contrast 90\% of the time. We note that the reported difficulty of these phenomena contradicts \cites{warstadt2019grammatical} conclusion that argument structure is one of the strongest domains for neural models. However, \citeauthor{warstadt2019grammatical} evaluate classifiers with supervision on CoLA, a large proportion of which is sentences related to argument structure. 

Finally, we caution against interpreting positive results on a general phenomenon in \BLiMP{} as proof of human-like knowledge. While it is unlikely that GPT-2 could reach human performance on the \textsc{subj.-verb agr.} paradigms without acquiring a concept of number marking that abstracts away from specific lexical items, it is difficult to rule out this possibility without accumulating different forms of evidence, for instance by testing how it generalizes to nonce words. We take the paradigms in \textsc{filler-gap} as a cautionary example (see Table \ref{tab:appendix}). There are four paradigms that assess a model's sensitivity to the syntactic requirements of complementizer \emph{that} versus a wh-word. We observe that all models more or less succeed when the unacceptable sentence lacks a necessary gap, but fail when it contains an illicit gap. 
These results suggest the models' ability to accurately detect a contrast in whether a gap is filled following a wh-word is not clearly based on a generalization about the relationship between that wh-word and its gap, as such a generalization should extend to the cases where the models currently fail to detect the correct contrast. More generally, conclusions about a model's knowledge of a particular grammatical concept can only be reached by considering several paradigms. 

\subsection{Shallow Predictors of Performance}\label{sec:shallow}

\begin{figure}[!t]
    \centering
    \includegraphics[trim=0 30 0 0, clip,width=0.5\textwidth]{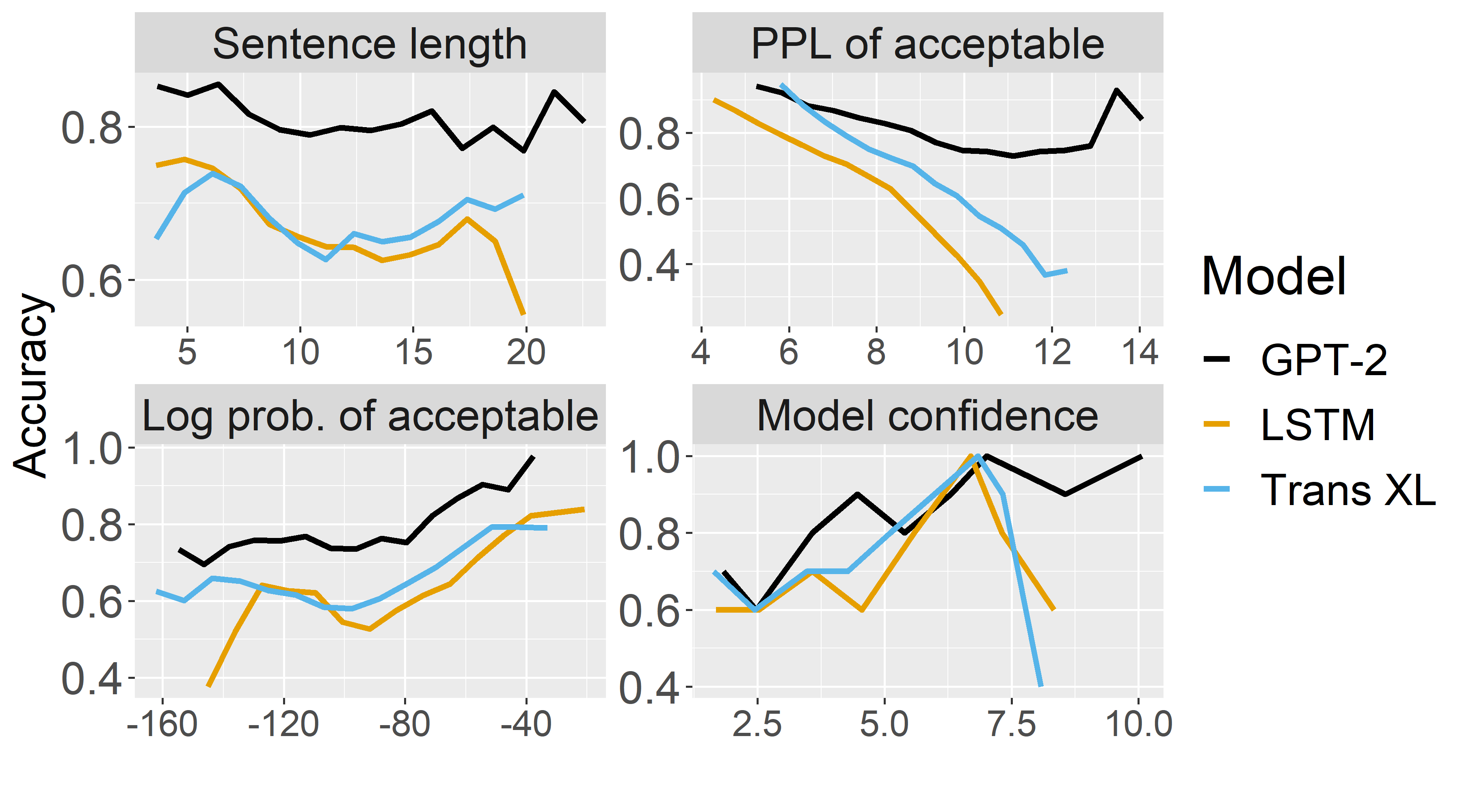}
    \caption{Models' performance on \BLiMP{} as a function of sentence length, perplexity, log probability of the acceptable sentence, and model confidence (calculated as $\left|\log P(S_1) - \log P(S_2)\right|$).}
    \label{fig:predict}
\end{figure}

We also ask what factors besides linguistic phenomena affect model accuracy.
Figure \ref{fig:predict} shows how sentence length, perplexity (which does not depend on length), the probability of the good sentence (which does depend on length), and confidence affect model performance. The effect of perplexity is much weaker for GPT-2 than for other models, which indicates it is probably more robust to sentences with non-stereotypical syntax or describing unlikely scenarios. GPT-2 is the only model where accuracy increases largely monotonically with confidence. A similar relationship holds between confidence and agreement in human acceptability judgments.

\subsection{Correlation of Model \& Human Performance}

We examine the extent to which models and humans succeed at detecting contrasts for the same linguistic phenomena. Figure \ref{tab:model_cor} shows the Pearson correlation between the four LMs and humans of their accuracies on the 67 paradigms. The neural models correlate moderately with humans, with GPT-2 correlating most strongly. The $n$-gram model's performance correlates with humans relatively weakly. Neural models correlate with each other more strongly, suggesting neural networks share some biases that are not human-like. Transformer-XL and LSTM's high correlation of 0.9 possibly reflects their similar training data.


\newcommand\hitems{5}   
\arrayrulecolor{white} 
\begin{figure}\centering
\scalebox{0.75}{
\small
\renewcommand{\arraystretch}{1.5}
\begin{tabular}{l*{\hitems}{|F}|}
5-gram & 0.34 & 0.39 & 0.58 & 0.59 & 1\\ \hhline{~*\hitems{|-}|}
LSTM & 0.49 & 0.63 & 0.9 & 1 & 0.59 \\ \hhline{~*\hitems{|-}|}
TXL & 0.48 & 0.68 & 1 & 0.9 & 0.58 \\ \hhline{~*\hitems{|-}|}
GPT-2  & 0.54 & 1 & 0.68 & 0.63 & 0.39\\ \hhline{~*\hitems{|-}|}
human & 1 & 0.54 & 0.48 & 0.49 & 0.34 \\ \hhline{~*\hitems{|-}|}
{} & \multicolumn{1}{p{4ex}}{human} & \multicolumn{1}{p{4ex}}{{GPT\Hyphdash*2}} & \multicolumn{1}{p{4ex}}{TXL} & \multicolumn{1}{p{4ex}}{LSTM} & \multicolumn{1}{l}{5-gram} \\ \hhline{~*\hitems{|-}|}
\end{tabular}
}
\caption{Heatmap showing the correlation between models' accuracies in each of the 67 paradigms.}\label{tab:model_cor}
\end{figure}

\section{Analysis}


\subsection{Long-Distance Dependencies}\label{sec:long-distance}

\begin{figure}
    \centering
    \includegraphics[width=0.42\textwidth]{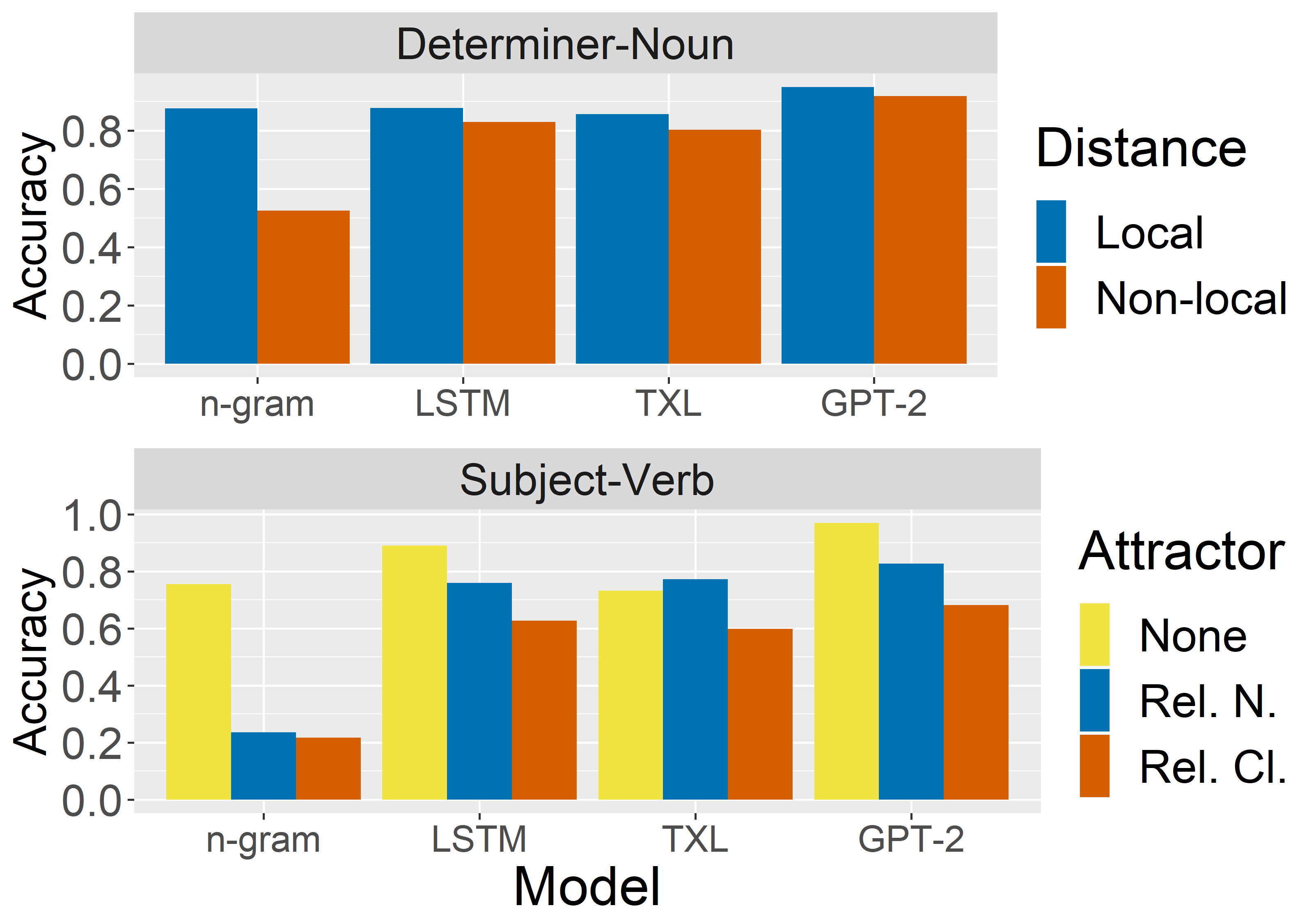}
    \caption{The effect of the locality of determiner-noun agreement (upper panel) and the type of agreement attractor (lower panel) on model performance.}
    \label{fig:long_distance}
\end{figure}

The presence of intervening material can lower the ability of humans to detect agreement dependencies \citep{bock1991broken}.
We study how intervening material affects the LMs' sensitivity to mismatches in agreement in \BLiMP{}. First, we test for sensitivity to determiner-noun agreement with and without an intervening adjective, as in Example \ref{eg:det_noun_adjective}. The results are plotted in Figure \ref{fig:long_distance}. The $n$-gram model is the most heavily impacted, performing on average $35$ points worse. This is unsurprising, since the bigram consisting of a determiner and noun is far more likely to be observed than the trigram of determiner, adjective, and noun. For the neural models, we find a weak but consistent effect, with all models performing on average between 5 and 3 points worse when there is an intervening adjective.

\ex.\label{eg:det_noun_adjective}
\a. Ron saw that man/*men.
\b. Ron saw that nice man/*men.

Second, we test for sensitivity to mismatches in subject-verb agreement when an \newterm{attractor} noun of the opposite number intervenes. We compare attractors in relative clauses \ref{eg:subject_verb_agr_distractor_b} and as part of a relational noun \ref{eg:subject_verb_agr_distractor_c}, following experiments by \citet{linzen2016assessing} and others. Again, we find that the $n$-gram model's performance is reduced significantly by this intervening material, suggesting the model is consistently misled by the presence of an attractor. All the neural models perform above chance with an attractor present, but GPT-2 and the LSTM perform 22 and 20 points worse when an attractor is present than when there is no attractor, while Transformer-XL's performance is reduced by only 5 points. Thus, we reproduce \citeauthor{linzen2016assessing}'s finding that attractors significantly reduce LSTM LMs' sensitivity to mismatches in agreement and find evidence that this holds true of some Transformer LMs as well. 

\ex.\label{eg:subject_verb_agr_distractor}
\a. The sisters bake/*bakes.
\b. The sisters who met Cheryl bake/*bakes.\label{eg:subject_verb_agr_distractor_b}
\b. The sisters of Cheryl bake/*bakes.\label{eg:subject_verb_agr_distractor_c}

\subsection{Regular vs. Irregular Agreement}\label{sec:regular-irregular}

In \textsc{det.-noun agr.} and \textsc{subj.-verb agr.}, we generate separate datasets for nouns with regular and irregular number marking, as in Example \ref{eg:regular_vs_irregular}. All else being equal, only models with access to sub-word-level information should make any distinction between regular and irregular morphology. 

\ex.\label{eg:regular_vs_irregular}
\a. Ron saw that nice kid/*kids. \hspace*{\fill}(regular)
\b. Ron saw that nice man/*men.
\hspace*{\fill}(irregular)

\begin{figure}
    \centering
    \includegraphics[trim=0 30 0 0, clip,width=0.42\textwidth]{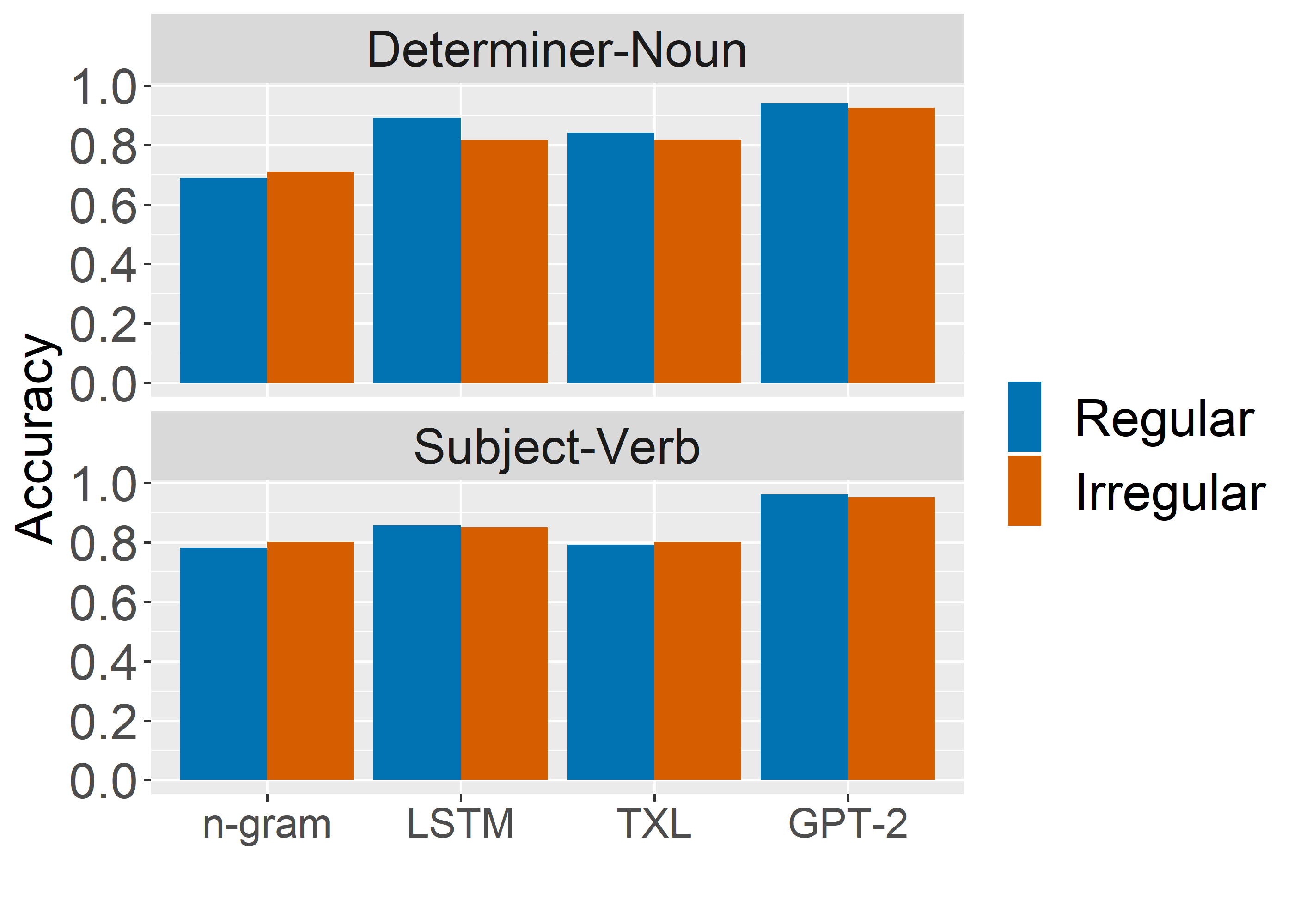}
    \caption{Models' performance on agreement phenomena between a determiner and noun and between a subject and verb, broken down by whether the noun/subject has a regular or irregular plural form}
    \label{fig:irregular}
\end{figure}

In fact, Figure \ref{fig:irregular} shows that the two sub-word-level models GPT-2 and Transformer-XL show little effect of irregular morphology: they perform less than $1.3$ points worse on irregulars than regulars. Their high overall performance suggests they robustly encode number features without relying on segmental cues.\footnote{The LSTM LM, which has word-level tokens, averages 5.2 points worse on the irregular paradigms. This effect is not due to morphology, but rather to the higher proportion of out-of-vocabulary items among the irregular nouns, which include many loanwords such as \emph{theses} and \emph{alumni}.}

\subsection{Training size and \BLiMP{} performance}\label{sec:data_learning_curve_experiment}

\begin{figure}[t]
    \centering
    \includegraphics[width=0.47\textwidth]{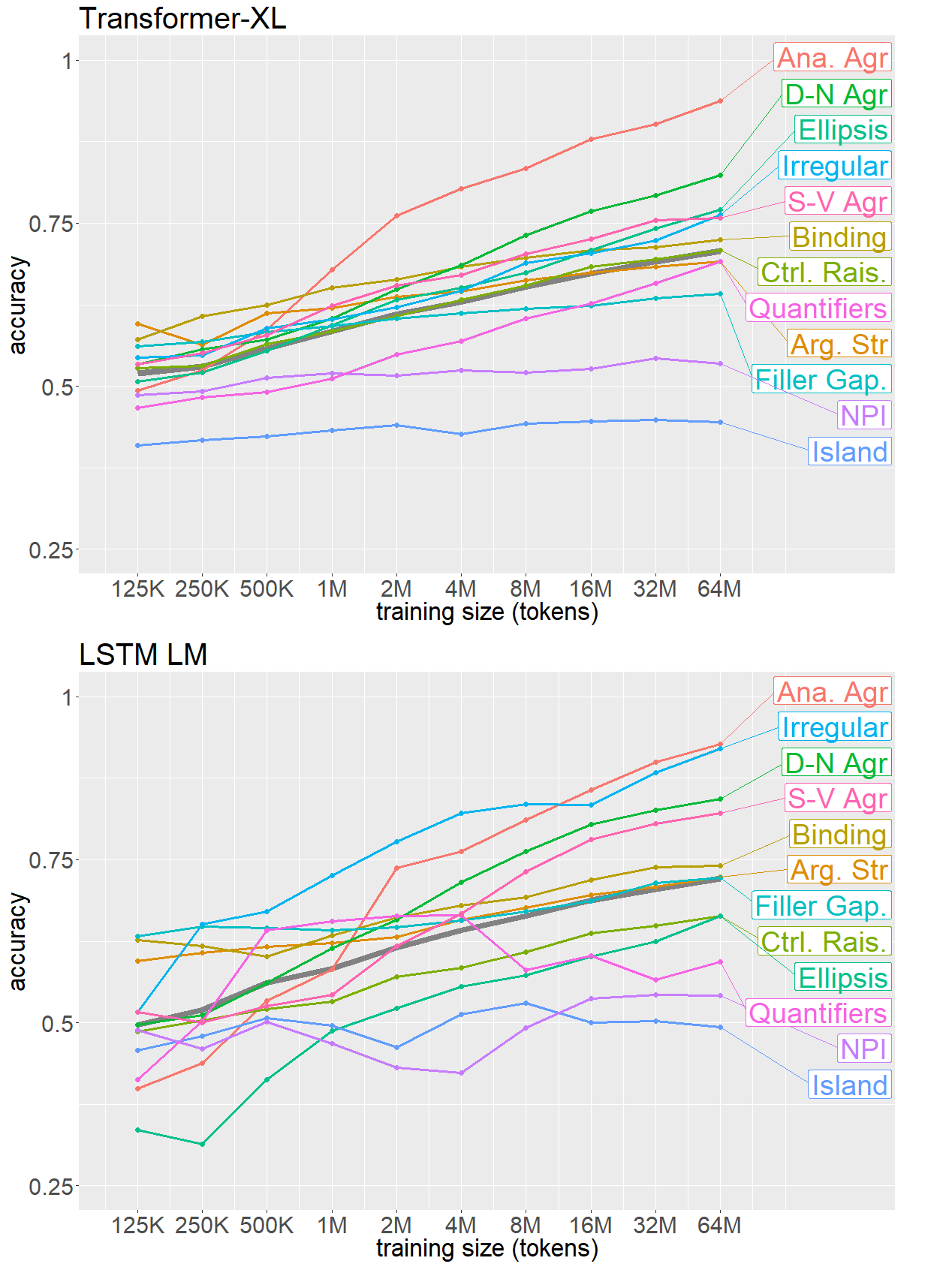}
    \caption{Transformer-XL (top) and LSTM LM (bottom) performance as a function of training size and phenomena in \BLiMP{}.
    The gray line shows the average across all phenomena.}
    \label{fig:LSTM_LM}
\end{figure}

We use \BLiMP{} to track how a model's representation of particular phenomena varies with the quantity of training data. Using different sized subsets of \cites{gulordava2019colorless} training data, we retrain the LSTM and Transformer-XL models and evaluate their performance on \BLiMP{}. Figure \ref{fig:LSTM_LM} shows that different phenomena have notably different learning curves across different training sizes even if the full model trained on 83M tokens achieved equivalent accuracy scores. For example, the LSTM model ultimately performs well on both \textsc{irregular} and \textsc{anaphor agr.}, but requires more training to reach this level of performance for \textsc{anaphor agr.}. These learning curve differences show how \BLiMP{} performance dissociates from perplexity on Wikipedia data, a standard measure of LM performance: although perplexity decreases with more training data,\footnote{Average perplexity on the \citet{gulordava2019colorless} test set: 595 at 0.125M, 212 at 1M, 92.8 at 8M, and 53 at 64M.} performance on different phenomena grows at varying rates.

We conjecture that there is a sigmoid relationship between the logarithm of training set size and \BLiMP{} performance which appears to be roughly linear at this scale. We conduct linear regression analyses to estimate the rate of increase in performance in relation to the logarithm (base 2) of dataset size. For the LSTM LM, best-fit lines for phenomena on which the model had the highest accuracy have the steepest slopes: \textsc{anaphor agr.} (0.0623), \textsc{det.-noun agr.} (0.0426), and \textsc{irregular} (0.039). We see the shallowest slopes on phenomena with the worst performance: \textsc{NPIs} (0.0078) and \textsc{islands} (0.0036).
For Transformer-XL, we observe a similar pattern: the steepest learning curves again belong to \textsc{anaphor agr.} (0.0545) and \textsc{det.-noun agr.} (0.0405), and the shallowest to \textsc{NPIs} (0.0055) and \textsc{Islands} (0.0039). Based on these values, we estimate that if log-linear improvement continues, the LSTM LM and Transformer-XL should require well over $10^{20}$ tokens of training data to achieve human-like performance on these hardest phenomena.

We also find that increasing model size (number of parameters) is unlikely to improve performance: We evaluate four pretrained versions of GPT-2 with 117M to 1558M parameters trained on WebText. All models have overall \BLiMP{} accuracy of $0.84\pm .01\%$, and standard deviation among the models on each of the 12 phenomena does not exceed $0.03$. This finding bolsters our earlier conclusion in \S \ref{sec:results} that amount of training data has the biggest impact on \BLiMP{} performance.

\subsection{Alternate Evaluation Methods}\label{sec:prefix}

There are several other methods one can use to measure an LM's preference between two minimally different sentences. So far, we have considered only the \newterm{full-sentence method}, advocated for by \citet{marvin2018targeted}, which compares LM likelihoods of full sentences. In a followup experiment, we use two \newterm{prefix methods}, each of which has appeared in related prior work, that evaluate a model's preferences by comparing its prediction at a key point of divergence between the sentences. Subsets of \BLiMP{} data are designed to be compatible with multiple methods, allowing us to conduct the first direct comparison. We find that all methods give broadly similar results when aggregating over a set of paradigms. We see no strong argument against evaluating solely using the full-sentence method, though some results diverge for specific paradigms.
     
\paragraph{One-Prefix Method}  In the \newterm{one-prefix method}, used by \citet{linzen2016assessing}, a pair of sentences share the same initial portion of a sentence, but differ in a critical word that make them differ in grammaticality (e.g., \emph{The cat \textbf{eats} mice} vs. \emph{The cat \textbf{eat} mice}). The model's prediction is correct if it assigns a higher probability to the grammatical token given the shared prefix.

\begin{figure}[t]
    \centering
    \includegraphics[width=0.5\textwidth]{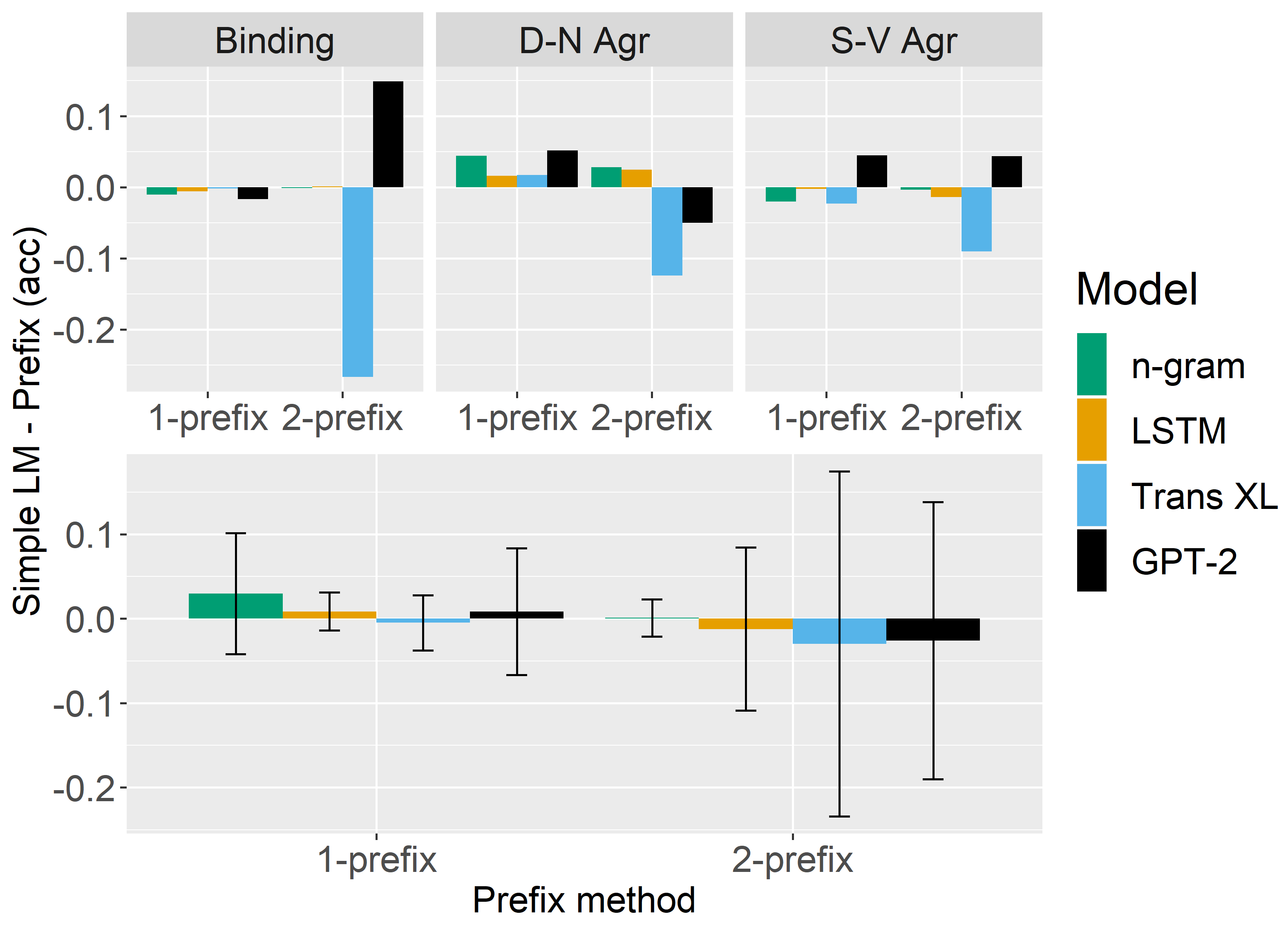}
    \caption{Comparison of models' performance on the simple LM method and the 1- and 2-prefix methods. The upper panels show results from three phenomena that are compatible with both 1-prefix and 2-prefix methods. The lower panel shows the averages and standard deviations across all phenomena.}
    \label{fig:prefix}
\end{figure}

\paragraph{Two-Prefix Method} In the \newterm{two-prefix method}, used by \citet{wilcox2019structural}, a pair of sentences differ in their initial string, and the grammaticality difference is only revealed when a shared critical word is included (e.g., \emph{The cat \textbf{eats} mice} vs.~\emph{The cats \textbf{eats} mice}). For these paradigms, we evaluate whether the model assigns a higher probability to the critical word conditioned on the grammatical prefix than on the ungrammatical prefix. 

The prefix methods differ from the full-sentence method in two key ways: (i) they require that the acceptability of the sentence be unambiguously predictable from the critical word, but not sooner, and (ii) they are not affected by predictions made by the LM following the critical word. These values do affect the full sentence method. For example, assuming that $P(\emph{are numerous}) \gg P(\emph{is numerous})$, a model could predict that \emph{The cats \textbf{are} numerous} is more likely than \emph{The cats \textbf{is} numerous} without correctly predicting that $P(\emph{are}|\emph{the cats}) > P(\emph{is}|\emph{the cats})$.
Using prefix probabilities allows us to exclude models' use of this additional information and evaluate how the models perform when they have just enough information to judge grammaticality.


Figure \ref{fig:prefix} shows that models have generally comparable accuracies across all three methods. However, there are some cases where we observe differences between these methods. For example, Transformer-XL performs much worse at \textsc{binding}, \textsc{det.-noun agr.}, and \textsc{subj.-verb agr.} in the simple LM method, suggesting that the probabilities Transformer-XL assigns to the irrelevant part at the end of the sentence very often overturn the observed preference based on probability up to the critical word. On the other hand, GPT-2 benefits from reading the whole sentence for \textsc{binding} phenomena, as its performance is better in the simple LM method than in the prefix method. 

We conclude that with a sufficiently diverse set of paradigms, the various metrics under consideration will give similar results. Thus, it is not problematic that \BLiMP{} relies only on the full-sentence method, and doing so allows \BLiMP{} to include many paradigms not compatible with either prefix method. Nonetheless, prefix methods are still valuable for detailed analysis or for studies making direct comparison to psycholinguistic theories \citep[e.g.][]{wilcox2018filler}.

\section{Conclusion \& Future Work} 

We have shown ways in which \BLiMP{} can be used as tool to gain evidence about both the overall and fine-grained linguistic knowledge of language models. Like the GLUE benchmark \citep{wang2018glue}, \BLiMP{} assigns a single overall score to an LM which summarizes its general sensitivity to minimal pair contrasts. It also provides a breakdown of LM performance by linguistic phenomenon, which can be used to draw more concrete conclusions about the kinds of grammatical features learned acquired by a given model. This kind of information is a linguistically motivated evaluation of LMs that can complement common metrics like perplexity.

Furthermore, the extent to which humans resemble data-driven learners like language models is debated in linguistics and cognitive science \citep[see e.g.,][]{chomsky1965aspects,reali2005uncovering}. In some domains, we may require the aid of innate knowledge to acquire phenomenon-specific knowledge resembling that tested in \BLiMP{}. By evaluating whether self-supervised learners like LMs acquire human-like grammatical acuity in a particular domain, we gather indirect evidence as to whether this phenomenon is a necessary component of humans' innate knowledge.



Another aim of \BLiMP{} is to serve as a guide for future work on the linguistic evaluation of LMs. It is particularly interesting to better understand those empirical domains where current LMs appear to acquire some relevant knowledge, but still fall short of human performance. The results from \BLiMP{} suggest that---in addition to relatively well-studied phenomena like filler-gap dependencies, NPIs, and binding---argument structure remains one area where there is much to uncover about what LMs learn. More generally, as language modeling techniques continue to improve, it will be useful to have large-scale tools like \BLiMP{} to efficiently track changes in what these models do and do not know about grammar.

\section*{Acknowledgments}

This material is based upon work supported by the National Science Foundation under Grant No. 1850208. Any opinions, findings, and conclusions or recommendations expressed in this material are those of the author(s) and do not necessarily reflect the views of the National Science Foundation. This project has also benefited from support to SB by Eric and Wendy Schmidt (made by recommendation of the Schmidt Futures program), by Samsung Research (under the project \textit{Improving Deep Learning using Latent Structure}), by Intuit, Inc., and by NVIDIA Corporation (with the  donation of a Titan V GPU). 

\bibliography{tacl2018}
\bibliographystyle{acl_natbib}

\onecolumn

\begin{multicols}{2}
\section*{Appendix}\footnotesize

The following contains examples from each of the 67 paradigms in \BLiMP{}.

\paragraph{\footnotesize Caveats} Some paradigms include non-transparent factors that may influence interpretation. We list here those factors that we are aware of:

\begin{itemize}[noitemsep,topsep=0pt]
    \item Several paradigms within \textsc{anaphor agreement} and \textsc{binding} rely on stereotyped gender assignment associated with names (e.g., \textit{Mary}). A model has to have at least a weak gender-name association in order to succeed on some paradigms in BLiMP. For example, we mark sentences like \emph{Mary hugged themselves} and \emph{Mary hugged himself} as unacceptable, and we never include possibilities like \emph{Mary hugged themself}.
    \item To isolate certain phenomena, we had to rely on acceptability contrasts present in mainstream US and UK English but absent in many other dialects. For example, some speakers would accept the sentence \emph{Suzy don't lie}, but we would mark this unacceptable based on mainstream US English judgments. BLiMP assesses models' knowledge of this specific dialect of English; in some cases it could \textit{penalize} models that conform to a different dialect. 
\end{itemize}

\columnbreak
\paragraph{\footnotesize How to read this table:}
\begin{itemize}[noitemsep,topsep=0pt]
    \item \textit{Phenomenon} refers to the linguistic phenomenon as noted in Table \ref{tab:phenomena table}. \textit{UID} refers to the unique identifier used in the released dataset.
    \item Model and human performance are reported as percent accuracy. `Human' uses the more conservative individual judgments (as opposed to majority vote, for which each paradigm would be either 100\% or 80\%).
    \item Each pair is marked for whether it is usable with a prefix method. All sentences are valid for the simple LM method.
    \item If a sentence has a checkmark (\checkmark) under the \textit{1pfx} column, the sentence can be used with the 1-prefix method in addition to the simple LM method. The bolded word is the critical word -- the probability of the two different critical words for the acceptable and unacceptable sentences can be compared based on the same `prefix'.
    \item If a sentence has a checkmark (\checkmark) under the \textit{2pfx} column, the sentence can be used with the 2-prefix method in addition to the simple LM method. The bolded word is the critical word -- the probability of that particular word can be compared based on the two different acceptable and unacceptable `prefixes'.
\end{itemize}

\end{multicols}

\renewcommand{\arraystretch}{1}
\arrayrulecolor{black}
\newcommand\phenomenon{0.07}
\newcommand\example{0.4}
\newcommand\UID{0.3}
\newcommand\prefixes{0.01}
\newcommand\dash{\cdashline{2-11}[0.5pt/2pt]}
\begin{table}[b!]\tiny

\begin{adjustbox}{width=\textwidth,center}
    \begin{tabular}{p{\phenomenon\textwidth}p{\UID\textwidth}GGGGGp{\example\textwidth}p{\example\textwidth}p{\prefixes\textwidth}p{\prefixes\textwidth}}
    \toprule
    Phenomenon 	&	 UID 	&	\multicolumn{1}{p{2.5ex}}{\rotatebox{\rotation}{5-gram}}	&	\multicolumn{1}{p{2.5ex}}{\rotatebox{\rotation}{LSTM}}	&	\multicolumn{1}{p{2.5ex}}{\rotatebox{\rotation}{TXL}}	&	\multicolumn{1}{p{2.5ex}}{\rotatebox{\rotation}{GPT-2}}	&	\multicolumn{1}{p{2.5ex}}{\rotatebox{\rotation}{Human}}	&	 Acceptable Example 	&	 Unacceptable Example 	&	 1pfx 	&	 2pfx 	\\	
    \midrule 																						
    \multirow{2}{\phenomenon\linewidth}{\textsc{Anaphor \mbox{agreement}}} 	&	 anaphor\allowbreak \_gender\allowbreak \_agreement 	&	44	&	88	&	91	&	99	&	96	&	 Katherine can't help \textbf{herself}. 	&	 Katherine can't help \textbf{himself}. 	&	 \checkmark 	&	 	\\	\dash
    {} 	&	 anaphor\allowbreak \_number\allowbreak \_agreement 	&	52	&	95	&	97	&	99	&	99	&	 Many teenagers were helping \textbf{themselves}. 	&	 Many teenagers were helping \textbf{herself}. 	&	 \checkmark 	&	 	\\	
    \midrule 																						
    \multirow{9}{\phenomenon\linewidth}{\textsc{Argument \mbox{structure}}} 	&	 animate\allowbreak \_subject\allowbreak \_passive 	&	70	&	72	&	74	&	77	&	86	&	 Amanda was respected by some \textbf{waitresses}. 	&	 Amanda was respected by some \textbf{picture}. 	&	 \checkmark 	&	 	\\	\dash
    {} 	&	 animate\allowbreak \_subject\allowbreak \_trans 	&	91	&	87	&	89	&	85	&	87	&	 Danielle \textbf{visited} Irene. 	&	 The eye \textbf{visited} Irene. 	&	 	&	 \checkmark 	\\	\dash
    {} 	&	 causative 	&	54	&	68	&	58	&	78	&	98	&	 Aaron breaks the glass. 	&	 Aaron appeared the glass. 	&	 	&	 	\\	\dash
    {} 	&	 drop\_argument 	&	72	&	79	&	70	&	81	&	87	&	 The Lutherans couldn't skate around. 	&	 The Lutherans couldn't disagree with. 	&	 	&	 	\\	\dash
    {} 	&	 inchoative 	&	51	&	65	&	54	&	66	&	82	&	 A screen was fading. 	&	 A screen was cleaning.  	&	 	&	 	\\	\dash
    {} 	&	 intransitive 	&	68	&	79	&	67	&	84	&	90	&	 Some glaciers are vaporizing. 	&	 Some glaciers are scaring.  	&	 	&	 	\\	\dash
    {} 	&	 passive\_1 	&	89	&	72	&	81	&	89	&	95	&	 Jeffrey's sons are insulted by Tina's supervisor. 	&	 Jeffrey's sons are smiled by Tina's supervisor.  	&	 	&	 	\\	\dash
    {} 	&	 passive\_2 	&	82	&	73	&	81	&	90	&	86	&	 Most cashiers are disliked. 	&	 Most cashiers are flirted.  	&	 	&	 	\\	\dash
    {} 	&	 transitive 	&	71	&	65	&	76	&	76	&	99	&	 A lot of actresses' nieces have toured \textbf{that} art gallery. 	&	 A lot of actresses' nieces have coped \textbf{that} art gallery. 	&	 	&	 \checkmark 	\\	
    \midrule																						
    \multirow{7}{\phenomenon\linewidth}{\textsc{Binding}} 	&	 principle\_A\allowbreak \_c\_command 	&	58	&	59	&	61	&	74	&	86	&	 A lot of actresses that thought about Alice healed \textbf{themselves}. 	&	 A lot of actresses that thought about Alice healed \textbf{herself}.  	&	 \checkmark 	&	 	\\	\dash
    {} 	&	 principle\_A\allowbreak \_case\_1 	&	100	&	100	&	100	&	100	&	98	&	 Tara thinks that \textbf{she} sounded like Wayne. 	&	 Tara thinks that \textbf{herself} sounded like Wayne.  	&	 \checkmark 	&	 	\\	\dash
    {} 	&	 principle\_A\allowbreak \_case\_2 	&	49	&	87	&	95	&	95	&	96	&	 Stacy imagines herself \textbf{praising} this actress. 	&	 Stacy imagines herself \textbf{praises} this actress.  	&	 \checkmark 	&	 	\\	\dash
    {} 	&	 principle\_A\allowbreak \_domain\_1 	&	95	&	98	&	99	&	99	&	95	&	 Carlos said that Lori helped \textbf{him}. 	&	 Carlos said that Lori helped \textbf{himself}.  	&	 \checkmark 	&	 	\\	\dash
    {} 	&	 principle\_A\allowbreak \_domain\_2 	&	56	&	68	&	70	&	78	&	75	&	 Mark imagines Erin might admire \textbf{herself}. 	&	 Mark imagines Erin might admire \textbf{himself}.  	&	 \checkmark 	&	 	\\	\dash
    {} 	&	 principle\_A\allowbreak \_domain\_3 	&	52	&	55	&	60	&	75	&	83	&	 Nancy could say every guy hides \textbf{himself}. 	&	 Every guy could say Nancy hides \textbf{himself}.  	&	 	&	 \checkmark 	\\	\dash
    {} 	&	 principle\_A\allowbreak \_reconstruction 	&	40	&	46	&	38	&	47	&	78	&	 It's herself who Karen criticized. 	&	 It's herself who criticized Karen. 	&	 	&	 	\\	
    \midrule																						
    \multirow{5}{\phenomenon\linewidth}{\textsc{Control/ \mbox{raising}}} 	&	 existential\_there\allowbreak \_object\allowbreak \_raising 	&	84	&	66	&	76	&	78	&	90	&	 William has declared \textbf{there} to be no guests getting fired. 	&	 William has obliged \textbf{there} to be no guests getting fired.  	&	 	&	 \checkmark 	\\	\dash
    {} 	&	 existential\_there\allowbreak \_subject\allowbreak \_raising 	&	77	&	80	&	79	&	91	&	88	&	 There was bound to be a fish escaping. 	&	 There was unable to be a fish escaping.  	&	 	&	 	\\	\dash
    {} 	&	 expletive\_it\allowbreak \_object\allowbreak \_raising 	&	72	&	63	&	72	&	79	&	86	&	 Regina wanted it to be \textbf{obvious} that Maria thought about Anna. 	&	 Regina forced it to be \textbf{obvious} that Maria thought about Anna.  	&	 	&	 \checkmark 	\\	\dash
    {} 	&	 tough\_vs\_raising\_1 	&	33	&	34	&	45	&	72	&	75	&	 Julia wasn't fun to talk to. 	&	 Julia wasn't unlikely to talk to.  	&	 	&	 	\\	\dash
    {} 	&	 tough\_vs\_raising\_2 	&	77	&	93	&	86	&	89	&	81	&	 Rachel was apt to talk to Alicia. 	&	 Rachel was exciting to talk to Alicia. 	&	 	&	 	\\	
    \midrule 																						
    \multirow{8}{\phenomenon\linewidth}{\textsc{Deter-miner-noun agr.}} 	&	 determiner\allowbreak \_noun\allowbreak \_agreement\_1 	&	88	&	92	&	92	&	99	&	96	&	 Craig explored that \textbf{grocery store}. 	&	 Craig explored that \textbf{grocery stores}.  	&	 \checkmark 	&	 	\\	\dash
    {} 	&	 determiner\allowbreak \_noun\allowbreak \_agreement\_2 	&	86	&	92	&	81	&	98	&	95	&	 Carl cures those \textbf{horses}. 	&	 Carl cures that \textbf{horses}.  	&	 	&	 \checkmark 	\\	\dash
    {} 	&	 determiner\allowbreak \_noun\allowbreak \_agreement\allowbreak \_irregular\_1 	&	53	&	76	&	81	&	93	&	92	&	 Phillip was lifting this \textbf{mouse}. 	&	 Phillip was lifting this \textbf{mice}.  	&	 \checkmark 	&	 	\\	\dash
    {} 	&	 determiner\allowbreak \_noun\allowbreak \_agreement\allowbreak \_irregular\_2 	&	55	&	83	&	77	&	94	&	85	&	 Those ladies walk through those \textbf{oases}. 	&	 Those ladies walk through that \textbf{oases}.  	&	 	&	 \checkmark 	\\	\dash
    {} 	&	 determiner\allowbreak \_noun\allowbreak \_agreement\allowbreak \_with\_adj\_1 	&	52	&	87	&	86	&	98	&	95	&	 Tracy praises those lucky \textbf{guys}. 	&	 Tracy praises those lucky \textbf{guys}.  	&	 \checkmark 	&	 	\\	\dash
    {} 	&	 determiner\allowbreak \_noun\allowbreak \_agreement\allowbreak \_with\_adj\_2 	&	50	&	86	&	78	&	95	&	96	&	 Some actors buy these gray \textbf{books}. 	&	 Some actors buy this gray \textbf{books}.  	&	 	&	 \checkmark 	\\	\dash
    {} 	&	 determiner\allowbreak \_noun\allowbreak \_agreement\allowbreak \_with\_adj\allowbreak \_irregular\_1 	&	53	&	76	&	81	&	93	&	94	&	 This person shouldn't criticize this upset \textbf{child}. 	&	 This person shouldn't criticize this upset \textbf{children}.  	&	 \checkmark 	&	 	\\	\dash
    {} 	&	 determiner\allowbreak \_noun\allowbreak \_agreement\allowbreak \_with\_adj\allowbreak \_irregular\_2 	&	55	&	83	&	77	&	94	&	85	&	 That adult has brought that purple \textbf{octopus}. 	&	 That adult has brought those purple \textbf{octopus}. 	&	 	&	 \checkmark 	\\	
    \midrule																						
    \multirow{2}{\phenomenon\linewidth}{\textsc{Ellipsis}} 	&	 ellipsis\_n\_bar\_1 	&	23	&	68	&	65	&	92	&	92	&	 Brad passed one big museum and Eva passed several. 	&	 Brad passed one museum and Eva passed several big.  	&	 	&	 	\\	\dash
    {} 	&	 ellipsis\_n\_bar\_2 	&	50	&	67	&	89	&	87	&	78	&	 Curtis's boss discussed four sons and Andrew discussed five sick sons. 	&	 Curtis's boss discussed four happy sons and Andrew discussed five sick. 	&	 	&	 	\\	
    \midrule 																						
    \multirow{7}{\phenomenon\linewidth}{\textsc{Filler gap}} 	&	 wh\_questions\allowbreak \_object\_gap 	&	53	&	79	&	61	&	84	&	85	&	 Joel discovered the vase that Patricia might take. 	&	 Joel discovered what Patricia might take the vase.  	&	 	&	 	\\	\dash
    {} 	&	 wh\_questions\allowbreak \_subject\_gap 	&	82	&	92	&	83	&	96	&	98	&	 Cheryl thought about some dog that upset Sandra. 	&	 Cheryl thought about who some dog upset Sandra.  	&	 	&	 	\\	\dash
    {} 	&	 wh\_questions\allowbreak \_subject\_gap\allowbreak \_long\_distance 	&	86	&	96	&	86	&	87	&	85	&	 Bruce knows that person that Dawn likes that argued about a lot of guys. 	&	 Bruce knows who that person that Dawn likes argued about a lot of guys.  	&	 	&	 	\\	\dash
    {} 	&	 wh\_vs\_that\allowbreak \_no\_gap 	&	83	&	97	&	86	&	97	&	97	&	 Danielle finds out that many organizations have alarmed Chad. 	&	 Danielle finds out who many organizations have alarmed Chad.  	&	 	&	 	\\	\dash
    {} 	&	 wh\_vs\_that\allowbreak \_no\_gap\allowbreak \_long\_distance 	&	81	&	97	&	91	&	94	&	92	&	 Christina forgot that all plays that win worry Dana. 	&	 Christina forgot who all plays that win worry Dana.  	&	 	&	 	\\	\dash
    {} 	&	 wh\_vs\_that\allowbreak \_with\_gap 	&	18	&	43	&	42	&	55	&	77	&	 Nina has learned who most men sound like. 	&	 Nina has learned that most men sound like.  	&	 	&	 	\\	\dash
    {} 	&	 wh\_vs\_that\allowbreak \_with\_gap\allowbreak \_long\_distance 	&	20	&	14	&	17	&	56	&	75	&	 Martin did find out what every cashier that shouldn't drink wore. 	&	 Martin did find out that every cashier that shouldn't drink wore. 	&	 	&	 	\\	
    \midrule																						
    \multirow{2}{\phenomenon\linewidth}{\textsc{Irregular forms}} 	&	 irregular\allowbreak \_past\allowbreak \_participle\allowbreak \_adjectives 	&	79	&	93	&	91	&	98	&	99	&	 The forgotten \textbf{newspaper article} was bad. 	&	 The forgot \textbf{newspaper article} was bad.  	&	 	&	 \checkmark 	\\	\dash
    {} 	&	 irregular\allowbreak \_past\allowbreak \_participle\allowbreak \_verbs 	&	80	&	85	&	66	&	86	&	95	&	 Edward \textbf{hid} the cats. 	&	 Edward \textbf{hidden} the cats. 	&	 \checkmark 	&	 	\\	
    \midrule 																						
    \multirow{8}{\phenomenon\linewidth}{\textsc{Island \mbox{effects}}} 	&	 adjunct\allowbreak \_island 	&	48	&	67	&	65	&	89	&	94	&	 Who has Colleen aggravated before kissing Judy? 	&	 Who has Colleen aggravated Judy before kissing?  	&	 	&	 	\\	\dash
    {} 	&	 complex\allowbreak \_NP\_ \allowbreak \_island 	&	50	&	47	&	58	&	72	&	80	&	 Who hadn't some driver who would fire Jennifer's colleague embarrassed? 	&	 Who hadn't Jennifer's colleague embarrassed some driver who would fire?  	&	 	&	 	\\	\dash
    {} 	&	 coordinate\allowbreak \_structure\allowbreak \_constraint\allowbreak \_complex\allowbreak \_left\allowbreak \_branch 	&	32	&	30	&	36	&	81	&	90	&	 What lights could Spain sell and \textbf{Andrea} discover? 	&	 What could Spain sell lights and \textbf{Andrea} discover?  	&	 	&	 \checkmark 	\\	\dash
    {} 	&	 coordinate\allowbreak \_structure\allowbreak \_constraint\allowbreak \_object\allowbreak \_extraction 	&	59	&	71	&	74	&	85	&	91	&	 Who will Elizabeth and Gregory cure? 	&	 Who will Elizabeth cure and Gregory?  	&	 	&	 	\\	\dash
    {} 	&	 left\allowbreak \_branch\allowbreak \_island\allowbreak \_echo\allowbreak \_question 	&	96	&	32	&	63	&	52	&	91	&	 David would cure what \textbf{snake}? 	&	 What would David cure \textbf{snake}?  	&	 	&	 \checkmark 	\\	\dash
    {} 	&	 left\allowbreak \_branch\allowbreak \_island\allowbreak \_simple\allowbreak \_question 	&	57	&	36	&	36	&	87	&	99	&	 Whose hat should Tonya wear? 	&	 Whose should Tonya wear hat?  	&	 	&	 	\\	\dash
    {} 	&	 sentential\allowbreak \_subject\allowbreak \_island 	&	61	&	43	&	37	&	36	&	61	&	 Who have many women's touring Spain embarrassed. 	&	 Who have many women's touring embarrassed Spain.  	&	 	&	 	\\	\dash
    {} 	&	 wh\_island 	&	56	&	47	&	20	&	79	&	73	&	 What could Alan discover \textbf{he} has run around? 	&	 What could Alan discover \textbf{who} has run around?  	&	 \checkmark 	&	 	\\	
    \midrule																						
    \multirow{7}{\phenomenon\linewidth}{\textsc{NPI \mbox{licensing}}} 	&	 matrix\allowbreak \_question\allowbreak \_npi\allowbreak \_licensor\allowbreak \_present 	&	1	&	2	&	1	&	65	&	98	&	 Should Monica \textbf{ever} grin? 	&	 Monica should \textbf{ever} grin.  	&	 	&	 \checkmark 	\\	\dash
    {} 	&	 npi\allowbreak \_present\_1 	&	47	&	54	&	61	&	65	&	83	&	 Even these trucks have \textbf{often} slowed. 	&	 Even these trucks have \textbf{ever} slowed.  	&	 \checkmark 	&	 	\\	\dash
    {} 	&	 npi\allowbreak \_present\_2 	&	47	&	54	&	48	&	64	&	98	&	 Many skateboards \textbf{also} roll. 	&	 Many skateboards \textbf{ever} roll.  	&	 \checkmark 	&	 	\\	\dash
    {} 	&	 only\allowbreak \_npi\allowbreak \_licensor\allowbreak \_present 	&	57	&	93	&	80	&	95	&	92	&	 Only Bill would \textbf{ever} complain. 	&	 Even Bill would \textbf{ever} complain.  	&	 	&	 \checkmark 	\\	\dash
    {} 	&	 only\allowbreak \_npi\allowbreak \_scope 	&	30	&	36	&	45	&	79	&	72	&	 Only those doctors who Karla respects \textbf{ever} conceal many snakes. 	&	 Those doctors who only Karla respects \textbf{ever} conceal many snakes.  	&	 	&	 \checkmark 	\\	\dash
    {} 	&	 sentential\allowbreak \_negation\allowbreak \_npi\allowbreak \_licensor\allowbreak \_present 	&	93	&	100	&	99	&	97	&	93	&	 Those banks had not \textbf{ever} lied. 	&	 Those banks had really \textbf{ever} lied.  	&	 	&	 \checkmark 	\\	\dash
    {} 	&	 sentential\allowbreak \_negation\allowbreak \_npi\allowbreak \_scope 	&	45	&	23	&	53	&	73	&	81	&	 Those turtles that are boring April could not \textbf{ever} break those couches. 	&	 Those turtles that are not boring April could \textbf{ever} break those couches. 	&	 	&	 \checkmark 	\\	
    \midrule																						
    \multirow{4}{\phenomenon\linewidth}{\textsc{Quantifiers}} 	&	 existential\allowbreak \_there\allowbreak \_quantifiers\_1 	&	91	&	96	&	94	&	100	&	94	&	 There aren't many lights darkening. 	&	 There aren't all lights darkening.  	&	 	&	 	\\	\dash
    {} 	&	 existential\allowbreak \_there\allowbreak \_quantifiers\_2 	&	62	&	16	&	14	&	42	&	76	&	 Each book is there disturbing Margaret. 	&	 There is each book disturbing Margaret.  	&	 	&	 	\\	\dash
    {} 	&	 superlative\allowbreak \_quantifiers\_1 	&	45	&	63	&	84	&	87	&	91	&	 No man has revealed more than five forks. 	&	 No man has revealed at least five forks.  	&	 	&	 	\\	\dash
    {} 	&	 superlative\allowbreak \_quantifiers\_2 	&	17	&	83	&	85	&	87	&	85	&	 An actor arrived at at \textbf{most} six lakes. 	&	 No actor arrived at at \textbf{most} six lakes.  	&	 	&	 \checkmark 	\\	
    \midrule																						
    \multirow{6}{\phenomenon\linewidth}{\textsc{Subject-verb agr.}} 	&	 distractor\allowbreak \_agreement\allowbreak \_relational\allowbreak \_noun 	&	24	&	76	&	77	&	80	&	81	&	 A sketch of lights \textbf{doesn't} appear. 	&	 A sketch of lights \textbf{don't} appear.  	&	 \checkmark 	&	 	\\	\dash
    {} 	&	 distractor\allowbreak \_agreement\allowbreak \_relative\allowbreak \_clause 	&	22	&	63	&	60	&	66	&	86	&	 Boys that aren't disturbing Natalie \textbf{suffer}. 	&	 Boys that aren't disturbing Natalie \textbf{suffers}.  	&	 	&	 \checkmark 	\\	\dash
    {} 	&	 irregular\allowbreak \_plural\allowbreak \_subject\allowbreak \_verb\allowbreak \_agreement\_1 	&	73	&	81	&	78	&	93	&	95	&	 This goose \textbf{isn't} bothering Edward. 	&	 This goose \textbf{weren't} bothering Edward. 	&	 \checkmark 	&	 	\\	\dash
    {} 	&	 irregular\allowbreak \_plural\allowbreak \_subject\allowbreak \_verb\allowbreak \_agreement\_2 	&	88	&	89	&	83	&	92	&	94	&	 The woman \textbf{cleans} every public park. 	&	 The women \textbf{cleans} every public park. 	&	 	&	 \checkmark 	\\	\dash
    {} 	&	 regular\allowbreak \_plural\allowbreak \_subject\allowbreak \_verb\allowbreak \_agreement\_1 	&	76	&	89	&	73	&	97	&	95	&	 Jeffrey \textbf{hasn't} criticized Donald. 	&	 Jeffrey \textbf{haven't} criticized Donald. 	&	 \checkmark 	&	 	\\	\dash
    {} 	&	 regular\allowbreak \_plural\allowbreak \_subject\allowbreak \_verb\allowbreak \_agreement\_2 	&	81	&	83	&	85	&	91	&	95	&	 The dress \textbf{crumples}. 	&	 The dresses \textbf{crumples}. 	&	 	&	 \checkmark 	\\	
        \bottomrule
    \end{tabular}
\end{adjustbox}
\caption{Examples of all 67 paradigms in \BLiMP{} along with model \% accuracy and estimated human agreement.}\label{tab:appendix}
\end{table}

\end{document}